\documentclass[runningheads]{llncs}
\usepackage[T1]{fontenc}
\usepackage{graphicx,verbatim}
\usepackage{amsmath,amssymb} 
\usepackage{xcolor} 
\usepackage{subcaption} 
\usepackage{algorithm2e} 
\RestyleAlgo{ruled}
\usepackage{booktabs} 
\usepackage{hyperref}
\usepackage[misc]{ifsym}

\begin{document}
\title{Lightweight Data-Free Denoising for Detail-Preserving Biomedical Image Restoration}
%

\author{Tomáš Chobola\inst{1,2}\orcidID{0009-0000-3272-9996}\textsuperscript{\Letter} \and
Julia A. Schnabel\inst{1,2,3}\orcidID{0000-0001-6107-3009} \and
Tingying Peng\inst{1,2}\orcidID{0000-0002-7881-1749}}
\index{Chobola, Tomáš}
\index{Schnabel, Julia}
\index{Peng, Tingying}
\authorrunning{T. Chobola et al.}
%
\institute{School of Computation, Information and Technology, Technical University of
Munich, Munich, Germany \and
Helmholtz AI, Helmholtz Munich - German Research Center for Environmental
Health, Neuherberg, Germany \and School of Biomedical Engineering and Imaging Sciences, King’s College London,
London, UK\\
\email{\{tomas.chobola,tingying.peng\}@helmholtz-munich.de}}
    
\maketitle              
\begin{abstract}
Current self-supervised denoising techniques achieve impressive results, yet their real-world application is frequently constrained by substantial computational and memory demands, necessitating a compromise between inference speed and reconstruction quality. In this paper, we present an ultra-lightweight model that addresses this challenge, achieving both fast denoising and high quality image restoration. Built upon the Noise2Noise training framework—which removes the reliance on clean reference images or explicit noise modeling—we introduce an innovative multistage denoising pipeline named Noise2Detail (N2D). During inference, this approach disrupts the spatial correlations of noise patterns to produce intermediate smooth structures, which are subsequently refined to recapture fine details directly from the noisy input. Extensive testing reveals that Noise2Detail surpasses existing dataset-free techniques in performance, while requiring only a fraction of the computational resources. This combination of efficiency, low computational cost, and data-free approach make it a valuable tool for biomedical imaging, overcoming the challenges of scarce clean training data—due to rare and complex imaging modalities—while enabling fast inference for practical use.

\keywords{Image denoising  \and Self-supervised \and Lightweight.}

\end{abstract}

\section{Introduction}

Image denoising is one of the fundamental tasks in computational imaging, particularly in medical and biological contexts, where it seeks to eliminate visual artifacts to enhance applications like medical diagnosis and microscopy imaging \cite{Li2022,Qu2024,Li2024}. State-of-the-art denoising methods, however, often grapple with achieving a balance between efficiency and performance in these fields. Typically, these approaches depend on large datasets of paired noisy and clean images to train image reconstruction models. While such techniques yield impressive results, their utility in biomedical imaging is curtailed by the challenges of acquiring such datasets which are not only costly and time-intensive but often impractical due to the inherent scarcity and complexity of obtaining noise-free biological samples, such as high-quality tissue scans or cellular images. Moreover, models trained on generic or unrelated datasets struggle to adapt to the unique noise distributions and intricate details of biomedical images, where even subtle deviations can compromise diagnostic accuracy. This is a significant limitation in clinical and biological research, where retaining fine details—such as cellular structures or pathological markers—is paramount for precise analysis and reliable diagnosis. Consequently, there is a pressing need for denoising techniques that are both data-efficient and adaptable, capable of delivering robust performance in medical and biological imaging without the harsh requirement of extensive paired training data.

To address these issues, several unsupervised denoising methods have been proposed as potential alternatives. Methods such as Noise2Noise \cite{n2n}, Noise2Void (N2V) \cite{krull2019noise2void}, Neighbor2Neighbor (Ne2Ne) \cite{neighbor2neighbor}, Zero-Shot Noise2Noise (ZS-N2N) \cite{zsn2n}, and others \cite{batson2019noise2self,quan2020self2self,ulyanov2018deep,wang2023noise2info} aim to utilize the intrinsic characteristics or patterns within the data, bypassing the requirement for noise-free reference images, or they harness the architecture of deep neural networks to extract low-level image details. However, these methods often suffer from significant limitations, such as high computational and memory requirements, suboptimal denoising performance, or reliance on prior knowledge of the noise distribution—information that is often difficult to obtain in real-world scenarios.

In this work, we build on Noise2Noise \cite{n2n} and Zero-Shot Noise2Noise \cite{zsn2n} to develop a lightweight denoising method termed Noise2Detail (N2D) that eliminates the need for noise model knowledge, extensive pre-training or clean reference datasets. Our approach uses a simple three-layer convolutional network designed for efficient execution on both GPUs and CPUs, enhancing its practicality in resource-constrained biomedical settings like clinical diagnostics or research labs. The restoration process begins with a simple network generating a \textit{partially restored image} after being trained on two downsampled versions of the noisy input, though background artifacts persist due to limited model capacity. Next, a multi-stage denoising step using pixel-shuffled images refines the background by breaking spatial noise correlations, albeit with a slight trade-off in  foreground sharpness. Finally, we re-use the network and fine-tune its weights to sharpen these critical foreground details, guided by the original noisy image. This balance of efficiency and performance makes it an effective tool for biomedical applications requiring rapid, precise denoising. We provide the source code here: \href{https://github.com/ctom2/noise2detail}{https://github.com/ctom2/noise2detail}. The contributions of this paper are as follows:
\begin{enumerate}
    \item \textbf{Lightweight Denoising Solution.}
    We propose a compact denoising model with just three convolutional layers, ensuring minimal computational cost. This enables rapid, high-quality image restoration, making it highly practical for biomedical imaging applications where computational resources are limited and fast inference is essential, such as in smart microscopy.
    \item \textbf{Data-Efficient Framework.} 
    Our three-step framework performs denoising without requiring clean reference data—an important advantage in biomedical imaging, where such data is often scarce or may not exist at all.
    \item \textbf{Diverse Biomedical Data Evaluation.} 
    We rigorously evaluate our method on CT scans with synthetic noise and a fluorescence microscopy dataset affected by real noise, demonstrating its ability to restore degraded images while preserving fine structural details.
\end{enumerate}

\section{Method}

\begin{figure}[t]
    \centering
    \includegraphics[width=.8\linewidth]{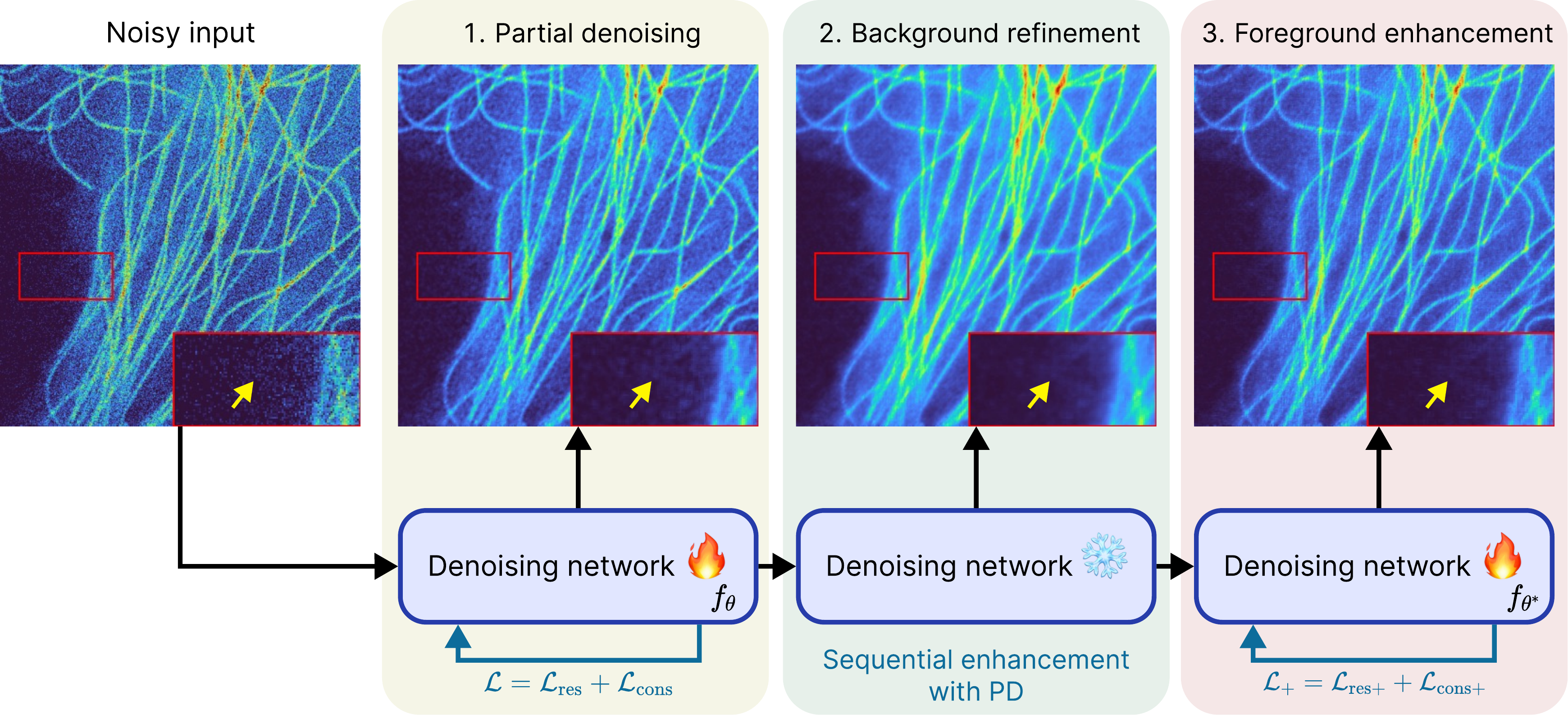}
    \caption{Overview of our proposed Noise2Detail denoising framework. Initially, $f_\theta$ is trained to denoise an image using a single-image Noise2Noise loss, but residual artifacts persist due to the limitations of the training (see the yellow arrow in the zoomed in insets). We then freeze the weights of the network, apply a sequential refinement to remove the background artifacts, and then retrain $f_{\theta^*}$ to enhance the foreground details, yielding the final denoised image.}
    \label{fig:architecture}
\end{figure}

\subsection{Revisiting Noise2Noise} 

In supervised image denoising, models $f_\theta$ are typically trained to map a noisy input $y$ to an estimate $f_\theta(y)$ of the clean image $x$, requiring paired noisy and clean data, where $y = x + e$ and $e$ is additive noise. Conversely, Noise2Noise~\cite{n2n} demonstrates that two noisy versions of the same image, $y_1 = x + e_1$ and $y_2 = x + e_2$, with $e_1$ and $e_2$ as independent noise instances, can substitute for noisy-clean pairs. The insight is that a model trained to predict $y_2$ from $y_1$ cannot learn the random noise but instead captures the consistent, correlated structure of the clean signal $x$, resulting in:
\begin{equation}
    \mathop{\arg \min}\limits_{\theta} \mathbb{E}\left[\lVert f_\theta(y_1) - x\rVert_2^2\right] = \mathop{\arg \min}\limits_{\theta} \mathbb{E}\left[\lVert f_\theta(y_1) - y_2\rVert_2^2\right].
\end{equation}
With infinite paired noisy samples, Noise2Noise theoretically matches the performance of training with clean data~\cite{n2n}, as the unpredictable noise is ignored in favor of the learnable signal. In practice, finite data causes a slight performance drop, and acquiring two pixel-aligned noisy images of the same scene is often impractical or costly, especially in applications like biomedical imaging, limiting its actual applications.

\subsection{Zero-Shot Denoising and Its Limitation}

To address the limitations of Noise2Noise training, ZS-N2N \cite{zsn2n} extends Ne2Ne \cite{neighbor2neighbor} to operate on single images through spatial downsampling. Given a noisy image $y\in\mathbb{R}^{H\times W\times C}$, two subsampled views $D_1(y),D_2(y)\in\mathbb{R}^{H/2\times W/2\times C}$ are generated via stride-2 convolutions with diagonal averaging kernels:

\begin{equation}
    k_1=
    \begin{bmatrix}
        0.5 & 0 \\
        0 & 0.5
    \end{bmatrix},
    k_2=
    \begin{bmatrix}
        0 & 0.5 \\
        0.5 & 0
    \end{bmatrix}.
\end{equation}
These kernels preserve local averages while creating pseudo-pairs for self-supervised training. The denoising network $f_\theta$ then learns a noise residual through a symmetric loss:
\begin{align}
\label{eq:residual_loss}
\begin{split}
    \mathcal{L}_\text{res}=\frac{1}{2}(&\lVert D_1(y) - f_\theta(D_1(y))-D_2(y)\rVert_2^2 \\
    + &\lVert D_2(y) - f_\theta(D_2(y))-D_1(y)\rVert_2^2),
\end{split}
\end{align}
To mitigate overfitting, the additional consistency loss enforces agreement between denoised and downsampled outputs:
\begin{align}
\label{eq:consistency_loss}
\begin{split}
    \mathcal{L}_\text{cons}=\frac{1}{2}(&\lVert D_1(y) - f_\theta(D_1(y))-D_2(y-f_\theta(y))\rVert_2^2 \\
    + &\lVert D_2(y) - f_\theta(D_2(y))-D_1(y-f_\theta(y))\rVert_2^2).
\end{split}
\end{align}
The total loss $\mathcal{L}=\mathcal{L}_\text{res}+\mathcal{L}_\text{cons}$ ensures stable training, and then the final denoised image can be obtained as $\bar{x}=y-f_\theta(y)$. However, averaging diagonal pixels poorly represents true noise distribution and retains spatial correlation of real noise \cite{zhou2020awgn}, causing artifacts and reduced performance. We use this as the first stage of our pipeline, with $f_\theta$ supporting later quality-enhancing steps.

\subsection{Refining Denoised Outputs with Pixel-Shuffle Techniques}

To mitigate the challenges above, we propose a two-stage refinement framework that breaks noise correlations while preserving authentic image structures from the input noisy image.

\subsubsection{Noise Correlation Elimination.} A key component of our proposed Noise2Detail method is the use of pixel-shuffle downsampling (PD) \cite{zhou2020awgn,lee2022ap} to disrupt spatial noise patterns. For stride $s$, PD decomposes $y$ into $s^2$ non-overlapping sub-images $\text{PD}_s(y) = \{ p_{s,i}(y) \}_{i=1}^{s^2}$, that then can be again rearrengend into the original image structure through an inverse operator $\text{PD}^{-1}$.

Thus, a naive and straighforward approach would be taking the noisy input image $y$, applying $\text{PD}$ to break the spatial correlations, denoise it using the trained model $f_\theta$, and then reshuffle the pixels back to their original positions. However, this approach introduces two critical issues: (1) The discontinuity between pixels in the denoised image produces grid-like artifacts as shown in \cite{pan2023random}, degrading the quality of the outputs. (2) While PD effectively breaks the spatial correlations between noisy pixels, it inherently also disrupts the spatial correlations of the signal pixels. This rearrangement can break the original structures, leading to over-smoothed results, as the model can no longer rely on local pixel relationships. 

\begin{figure}[t]
    \centering
    \includegraphics[width=.7\linewidth]{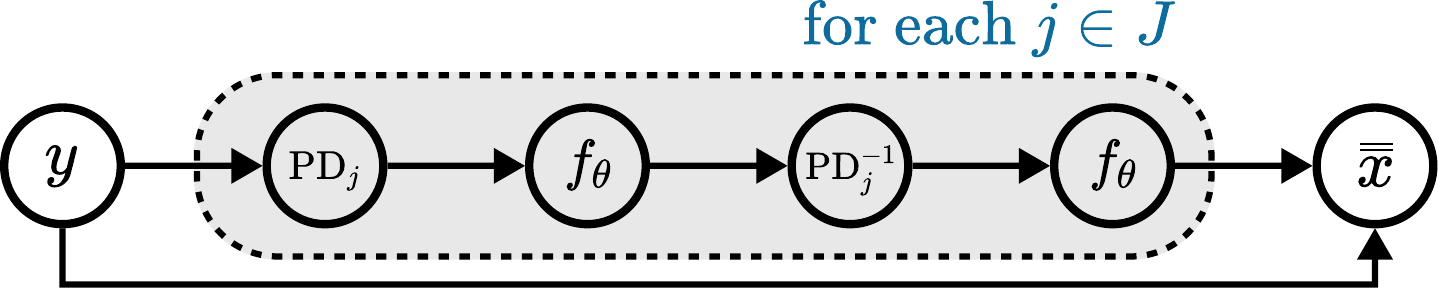}
    \caption{Illustration of the sequential enhancement process with a trained network $f_\theta$. For every $j \in J$, $f_\theta$ is applied twice—initially to denoise the pixel-shuffled downsampled image, and then to correct the grid-like artifacts caused by disrupted spatial correlations when reconstructing the original pixel layout.}
    \label{fig:step2}
\end{figure}

To address those challenges, we introduce a sequential denoising refinement process. After computing the initial PD-denoised estimate $\tilde{x}_j$, we leverage the pre-trained model $f_\theta$ to further denoise the unshuffled image as follows:
\begin{align}
    \tilde{x}_j &= \text{PD}^{-1}_j\left(\text{PD}_j(y) - f_\theta(\text{PD}_j(y))\right), \\
    \hat{x}_j &= \tilde{x}_j - f_\theta(\tilde{x}_j).
\end{align}
However, while this refinement step mitigates unwanted artifacts in the denoised image, the use of PD breaks the spatial correlations between pixels, resulting in a smoothing effect that blurs fine details and sometimes eliminates them completely. This happens because downsampling reduces the number of pixels available for the model to predict detailed structures, limiting its ability to reconstruct the full resolution of intricate features present in the original image. To overcome this, we blend the original partially denoised output $\bar{x}$ with the refined outputs $\hat{x}_j$, effectively merging preserved structural details with smoother regions into a unified representation:
\begin{equation}
    \bar{\bar{x}} = \frac{1}{1 + |J|} \left( \bar{x} + \sum_{j \in J} \hat{x}_j \right).
\end{equation}
We illustrate the refinement process in Figure \ref{fig:step2}.

\subsubsection{Detail Recovery.} To restore lost foreground details, we introduce a detail enhancement process. We repurpose the network $f_\theta$ and retrain it with a modified residual and consistency loss, as defined in Equations \ref{eq:res_plus} and \ref{eq:cons_plus}, respectively. Here, the network $f_{\theta^*}$ uses the partially denoised image $\bar{\bar{x}}$ as input and targets the original noisy image $y$. Since the network cannot predict the random noise in $y$ from the denoised $\bar{\bar{x}}$, it focuses on refining the structural details, effectively enhancing the foreground features.
\begin{align}
\label{eq:res_plus}
\begin{split}
    \mathcal{L}_\text{res+}=\frac{1}{2}(&\lVert f_{\theta^*}(D_1(\bar{\bar{x}}))-D_2(y)\rVert_2^2 \\
    + &\lVert f_{\theta^*}(D_2(\bar{\bar{x}}))-D_1(y)\rVert_2^2),
\end{split}
\end{align}
\begin{align}
\label{eq:cons_plus}
\begin{split}
    \mathcal{L}_\text{cons+}=\frac{1}{2}(&\lVert f_{\theta^*}(D_1(\bar{\bar{x}}))-D_2(f_{\theta^*}(\bar{\bar{x}}))\rVert_2^2 \\
    +&\lVert f_{\theta^*}(D_2(\bar{\bar{x}}))-D_1(f_{\theta^*}(\bar{\bar{x}}))\rVert_2^2).
\end{split}
\end{align}
With the trained network $f_{\theta^*}$, we then obtain the final denoised image as follows, $\dot{x}=f_{\theta^*}(\bar{\bar{x}})$. The complete denoising framework is visualized in Figure \ref{fig:architecture} with an image from BioSR dataset \cite{Qiao2020}.

\section{Experiments and Results}

\subsection{Datasets and Performance Comparison}

We conduct experiments on data with synthetic and real noise. Using a 33-slice CT scan of a spleen from the MSD dataset \cite{Antonelli2022}, we apply Gaussian and Poisson noise. For real-world data, we use the FMD testing set \cite{zhang2019poisson} consisting of 48 images, featuring noisy fluorescence microscopy images of cells, zebrafish, and mouse brain tissues, with clean ground-truth images averaged from 50 noisy observations. We compare our method to state-of-the-art self-supervised denoising methods: DIP \cite{ulyanov2018deep}, N2V \cite{krull2019noise2void}, Ne2Ne \cite{neighbor2neighbor}, N2S \cite{batson2019noise2self}, and ZS-N2N \cite{zsn2n}, using original training setups except for Ne2Ne and N2S, where we apply single-image adaptations from \cite{lequyer2021noise2fast} to make the comparisons fair.

\subsection{Implementation Details}

Our model $f$ uses a compact design with just three convolutional layers. The first two transform input channels into a 48-dimensional feature space, and the final layer converts these back to image space. With only about 22,000 parameters, it is far smaller and more efficient than existing methods, which often have millions of parameters requiring high computational costs. We train parameters $\theta$ and $\theta^*$ for 2,000 iterations at a $1 \times 10^{-3}$ learning rate. For refinement, we use $j \in \{2, 4\}$, optimized based on our ablation study in Subsection \ref{subsection:ablation}. Results and inference times come from testing on an NVIDIA A100 GPU.

\subsection{Ablation Study}\label{subsection:ablation}

To analyze the impact of different combinations of PD-processed images in the refinement step on denoising a spleen CT scan degraded by synthetic Poisson noise, and to justify hyperparameter choices, we conducted an ablation study. The results, detailed in Table \ref{tab:synthetic_comparison_onehot}, demonstrate that incorporating images processed with broken spatial correlation significantly improves the performance of our proposed framework, thus supporting our design approach. Specifically, the highest performance is achieved when combining images for $j\in\{2,4\}$, while including additional image for $j\in\{2,4,8\}$ leads to a performance drop due to over-smoothing, which reduces detail visibility.

\begin{table}
\centering
\caption{Performance comparison based on PSNR with indicators for included variables under synthetic Poisson noise $\lambda \in [10,50]$.}
\label{tab:synthetic_comparison_onehot}
\begin{tabular}{ccccc}
\toprule
\(\bar{x}\) & \(\hat{x}_2\) & \(\hat{x}_4\) & \(\hat{x}_8\) & PSNR$\uparrow$ (dB) \\
\midrule
\checkmark &  &  &  & 30.75 \\
\checkmark & \checkmark &  &  & 31.48 \\
\checkmark & \checkmark & \checkmark &  & $\mathbf{31.56}$ \\
\checkmark & \checkmark & \checkmark & \checkmark & 31.23 \\
\bottomrule
\end{tabular}
\end{table}

\subsection{Results}

\begin{figure}[t]
\centering
\begin{subfigure}{.162\textwidth}
    \centering
    \includegraphics[width=\linewidth]{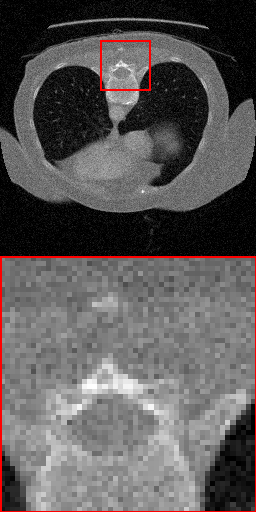}
    \caption{Input}
\end{subfigure}%
\hfill
\begin{subfigure}{.162\textwidth}
    \centering
    \includegraphics[width=\linewidth]{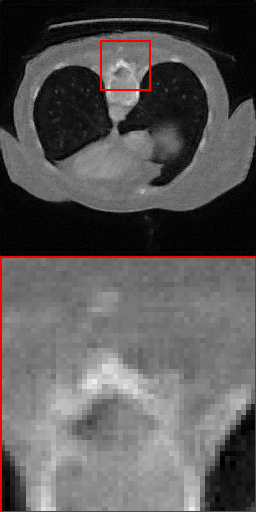}
    \caption{N2V}
\end{subfigure}%
\hfill
\begin{subfigure}{.162\textwidth}
    \centering
    \includegraphics[width=\linewidth]{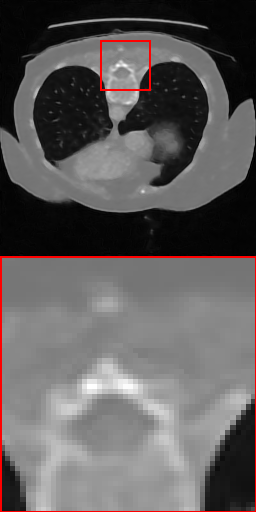}
    \caption{N2S}
\end{subfigure}%
\hfill
\begin{subfigure}{.162\textwidth}
    \centering
    \includegraphics[width=\linewidth]{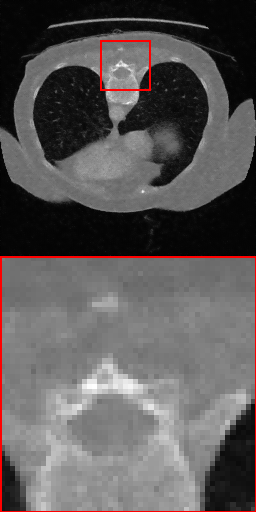}
    \caption{ZS-N2N}
\end{subfigure}%
\hfill
\begin{subfigure}{.162\textwidth}
    \centering
    \includegraphics[width=\linewidth]{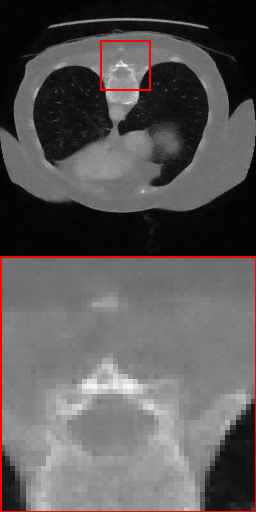}
    \caption{Ours}
\end{subfigure}%
\hfill
\begin{subfigure}{.162\textwidth}
    \centering
    \includegraphics[width=\linewidth]{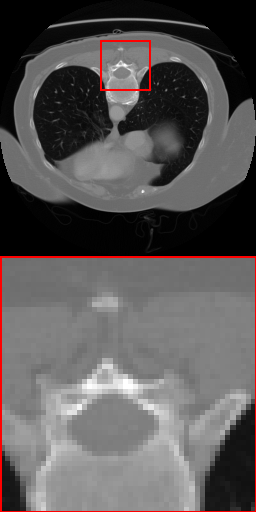}
    \caption{GT}
\end{subfigure}%
\caption{Our method is the only one that maintains the fidelity of the foreground while keeping the background smooth without artifacts.}
\label{fig:ct_comparison}
\end{figure}

\begin{figure}[t]
\centering
\begin{subfigure}{.198\textwidth}
    \centering
    \includegraphics[width=\linewidth]{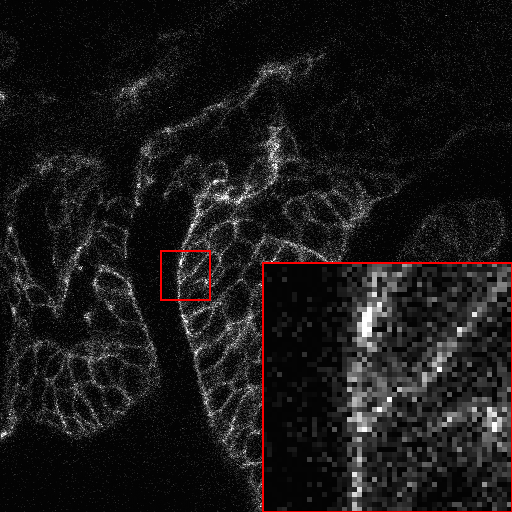}
    \caption{Input}
\end{subfigure}%
\hfill
\begin{subfigure}{.198\textwidth}
    \centering
    \includegraphics[width=\linewidth]{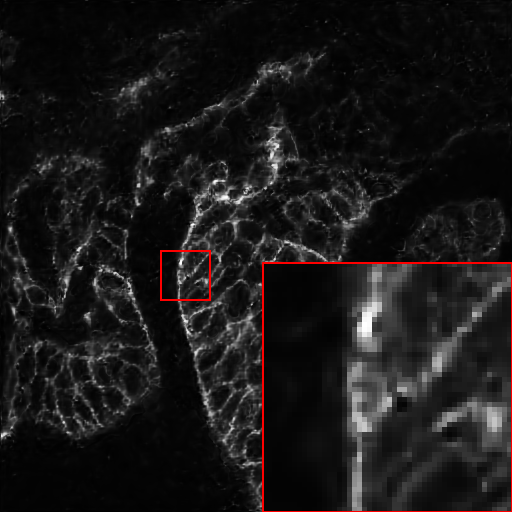}
    \caption{DIP}
\end{subfigure}%
\hfill
\begin{subfigure}{.198\textwidth}
    \centering
    \includegraphics[width=\linewidth]{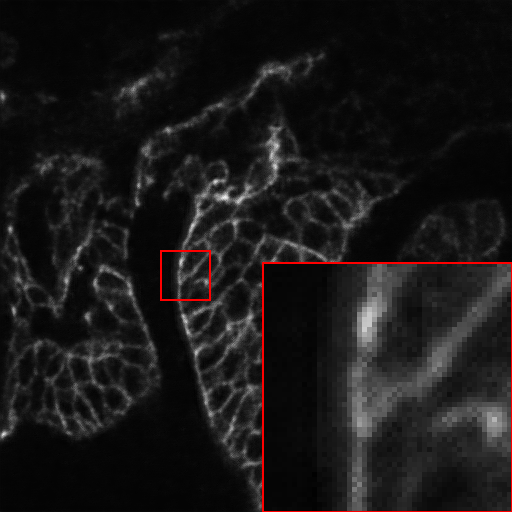}
    \caption{N2V}
\end{subfigure}%
\hfill
\begin{subfigure}{.198\textwidth}
    \centering
    \includegraphics[width=\linewidth]{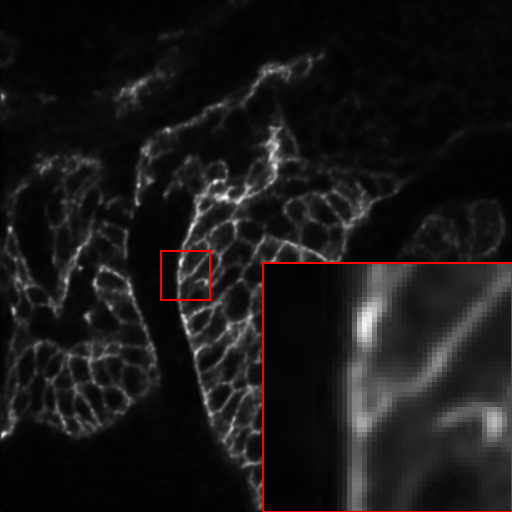}
    \caption{N2S}
\end{subfigure}%
\hfill
\begin{subfigure}{.198\textwidth}
    \centering
    \includegraphics[width=\linewidth]{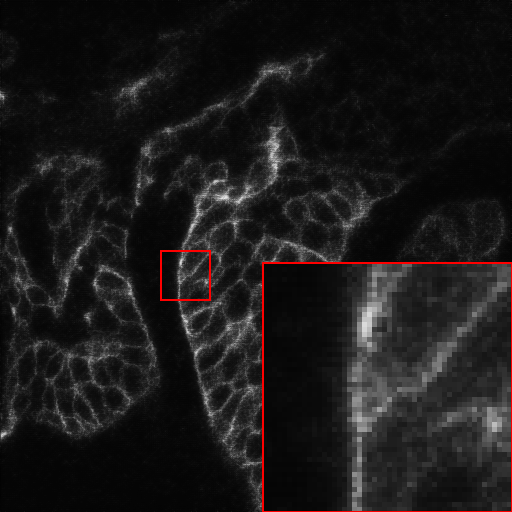}
    \caption{Ours}
\end{subfigure}%
\caption{Examples of restoration given a noisy image of a zebrafish from a confocal microscope taken from FMD \cite{zhang2019poisson}. N2V and N2S oversmooth and blur the details while our method maintains the structural fidelity.}
\label{fig:fmd_comparison}
\end{figure}

Our quantitative analysis of results for removing synthetic and real microscopy noise is shown in Tables \ref{tab:synthetic_comparison} and \ref{tab:fmd_comparison}. Table \ref{tab:synthetic_comparison} lists the average inference times for $512\times 512$ pixel images from the FMD dataset. Our proposed approach N2D achieves the best performance for synthetic noise removal and second best performance for real microscopy noise, outperformed only by N2S. However, the inference time of our method is significantly lower than N2S. This combination of strong performance and low computational cost emphasizes the strength of our method, positioning it as a competitive alternative to larger and expensive methods. Figures \ref{fig:ct_comparison} and \ref{fig:fmd_comparison} visualize the denoising results, demonstrating that our method uniquely delivers high-quality results without hallucinating background artifacts, and maintaining high levels of detail fidelity.

\begin{table}
\centering
\caption{Performance comparison across different synthetic noise levels with PSNR$\uparrow$ (dB). Best result is in \textbf{bold}, second best result is \underline{underlined}.}
\label{tab:synthetic_comparison}
\begin{tabular}{lcccccc}
\toprule
Noise         & DIP & N2V & Ne2Ne & N2S & ZS-N2N & \textbf{N2D} (ours) \\
\midrule
Gaussian $\sigma=25$                 & 27.31 & 27.59 & 19.34 & \underline{29.18} & 28.97 & $\mathbf{29.36}$ \\
Gaussian $\sigma\in[10,50]$          & 26.62 & 26.94 & 18.87 & \underline{28.81} & 28.47 & $\mathbf{28.84}$ \\
Poisson $\lambda\in[10,50]$          & 27.08 & 28.65 & 19.97 & \underline{31.50} & 31.28 & $\mathbf{31.56}$ \\
\midrule
Average                              & 27.00 & 27.73 & 19.40 & \underline{29.83} & 29.57 & $\mathbf{29.92}$ \\
\bottomrule
\end{tabular}
\end{table}

\begin{table}[h!]
\centering
\caption{Performance comparison across subsets of FMD \cite{zhang2019poisson} with PSNR$\uparrow$ (dB), and average inference times per image. Best result is in \textbf{bold}, second best result is \underline{underlined}.}
\label{tab:fmd_comparison}
\begin{tabular}{lcccccc}
\toprule
Subset         & DIP & N2V & Ne2Ne & N2S & ZS-N2N & \textbf{N2D} (ours) \\
\midrule
Confocal\_BPAE          & 31.07        & 33.58       &  14.93  &  \underline{34.92}  & 34.82       & $\mathbf{35.35}$         \\
Confocal\_FISH          & 23.99        & 30.26       &  14.82  &  $\mathbf{31.86}$  & 30.46          & \underline{30.80}         \\
Confocal\_MICE          & 29.60        & 35.06       &  11.98  &  $\mathbf{36.80}$  & 35.95          & \underline{36.22}         \\
TwoPhoton\_BPAE         & 27.35        & 32.19       &  16.79  &  $\mathbf{33.59}$  & 32.69          & \underline{32.95}         \\
TwoPhoton\_MICE         & 25.41        & 32.59       &  16.62  &  $\mathbf{33.33}$  & 32.71          & \underline{33.14}         \\
WideField\_BPAE         & 28.40        & 25.98       &  17.16  &  \underline{25.54}  & 25.83          & $\mathbf{26.20}$         \\
\midrule
Average    & 27.64        & 31.61       &  15.38  &  $\mathbf{32.67}$  & 32.08          & \underline{32.44}         \\
\midrule
Inference time (s) & 82 & 42 & 413 & 1250 & 13 & 24 \\
\bottomrule
\end{tabular}
\end{table}

\section{Conclusion}

In this work, we present Noise2Detail, a lightweight denoising framework designed for biomedical imaging, where clean data is scarce and the use of large models are often not feasible. Built on Noise2Noise, our three-layer network employs a sequential process of refinement, background correction, and detail enhancement. It restores images efficiently without requiring noise model assumptions or extensive datasets. Evaluations on medical and biological datasets confirm its ability to preserve details and remove noise effectively. This scalable method supports rapid denoising, enhancing biomedical research applications.

\begin{credits}
\subsubsection{\ackname} Tomáš Chobola is supported by the Helmholtz Association under the joint research school "Munich School for Data Science - MUDS".

\subsubsection{\discintname} The authors have no competing interests to declare that are relevant to the content of this article.
\end{credits}

%
%
%
%
\bibliographystyle{splncs04}

\end{document}